%% file: main.tex
\documentclass[letterpaper,twocolumn,fleqn]{article} 

\usepackage{ist}
\usepackage{xcolor}

\usepackage{cite}
\usepackage{times}
\usepackage{epsfig}
\usepackage{graphicx}
\usepackage{amsmath}
\usepackage{amssymb}
\usepackage{textcomp}
\usepackage{multirow}
\usepackage{adjustbox}
\usepackage{indentfirst}
\usepackage{multirow}
\usepackage{colortbl}
\usepackage{hhline}
\usepackage{subcaption}
\usepackage{enumitem}
\usepackage{verbatim}
\usepackage{tabularx}
\usepackage{booktabs}
\usepackage{float}

\usepackage[symbol]{footmisc}

\usepackage{gensymb}
\usepackage{flushend}
\usepackage{xcolor}
\usepackage{xurl}
\usepackage[colorlinks]{hyperref} 
\hypersetup{
    colorlinks=true,
    breaklinks=true, 
    urlcolor= blue,              
    linkcolor= red, 
    bookmarksopen=false, 
    filecolor=black, 
    citecolor=green, 
    linkbordercolor=blue
}
\usepackage{dirtytalk}

\title{Neural Rendering based Urban Scene Reconstruction for \\ Autonomous Driving}

\author{Shihao Shen$^{\ 1}$, Louis Kerofsky$^{\ 1}$, Varun Ravi Kumar$^{\ 1}$ and Senthil Yogamani$^{\ 2}$   \\ 
$^{1}$Qualcomm Technologies, Inc., San Diego, California, U.S. \\
$^{2}$Automated Driving, QT Technologies Ireland Limited.}

\date{} 

\begin{document} 

\maketitle 

\thispagestyle{empty} 
\input{include/abstract.tex}
\input{include/intro.tex}

\input{include/method.tex}

\input{include/experiment.tex} 

\input{include/conclusions.tex}

{\small
\bibliographystyle{ieeetr}
\bibliography{references}
}


\input{include/bio}

\end{document}

%% file: include/abstract.tex
\begin{abstract}
Dense 3D reconstruction has many applications in automated driving including automated annotation validation, multimodal data augmentation, providing ground truth annotations for systems lacking LiDAR, as well as enhancing auto-labeling accuracy. LiDAR provides highly accurate but sparse depth, whereas camera images enable estimation of dense depth but noisy particularly at long ranges. In this paper, we harness the strengths of both sensors and propose a multimodal 3D scene reconstruction using a framework combining neural implicit surfaces and radiance fields. In particular, our method estimates dense and accurate 3D structures and creates an implicit map representation based on signed distance fields, which can be further rendered into RGB images, and depth maps. A mesh can be extracted from the learned signed distance field and culled based on occlusion. Dynamic objects are efficiently filtered on the fly during sampling using 3D object detection models. We demonstrate qualitative and quantitative results on challenging automotive scenes.
\end{abstract}

%% file: include/intro.tex
\section{INTRODUCTION} 
\label{sec:Intro}

Advancement in the field of computer vision has enabled the rapid development of perception systems for autonomous vehicles (AV) in recent years. Deep learning in particular has accelerated this progress by achieving rapid advancement in various perception tasks including object detection~\cite{ mohapatra2021bevdetnet, rashed2020fisheyeyolo, rashed2021generalized}, semantic segmentation~\cite{chennupati2019auxnet, rashed2019motion, briot2018analysis}, depth prediction~\cite{thesis_varun, kumar2021svdistnet, kumar2018monocular, kumar2020fisheyedistancenet, kumar2021syndistnet, kumar2021fisheyedistancenet++, sekkat2022synwoodscape}, adverse weather detection~\cite{dhananjaya2021weather, uricar2019desoiling, das2020tiledsoilingnet, shen2023optical}, moving object detection~\cite{siam2018modnet, mohamed2021monocular, ramzy2019rst}, SLAM~\cite{tripathi2020trained, yahiaoui2019overview, kumar2023surround}, multi-task learning~\cite{leang2020dynamic, kumar2021omnidet, rashed2019optical} and sensor fusion~\cite{dasgupta2022spatio, uricar2019challenges, eising2021near}. However, dense scene construction of urban scenes is relatively less mature due to a variety of challenges. Firstly, it is challenging to reconstruct a spatially consistent dense structure map over time due to odometry errors. Secondly, there are many moving objects in urban scenes which hinders the 3D reconstruction process. Finally, urban scenes have fine 3D structures like curb and poles which are challenging to estimate.

3D scene reconstruction, refers to the creation of three-dimensional models from available data modalities, e.g. a set of images. Traditionally, 3D scenes are represented by point clouds, voxels, or meshes that are explicit and discrete. Recent Neural Radiance Fields (NeRF) models are able to represent a continuous scene implicitly through Multi-Layer Perceptrons (MLPs) that learn the scene geometry and appearance simultaneously. Since NeRF was first introduced by Mildenhall et al.~\cite{mildenhall2021nerf}, it has been widely explored and adapted into variants for a variety of applications, including city reconstruction~\cite{turki2022mega}, image processing~\cite{wang2022nerf}, generative AI~\cite{poole2022dreamfusion} to point out a few. This paper aims to tackle the application of reconstructing unconstrained urban scenes from monocular recordings. 

\begin{figure}[t!]
\centering
\includegraphics[width=\columnwidth]{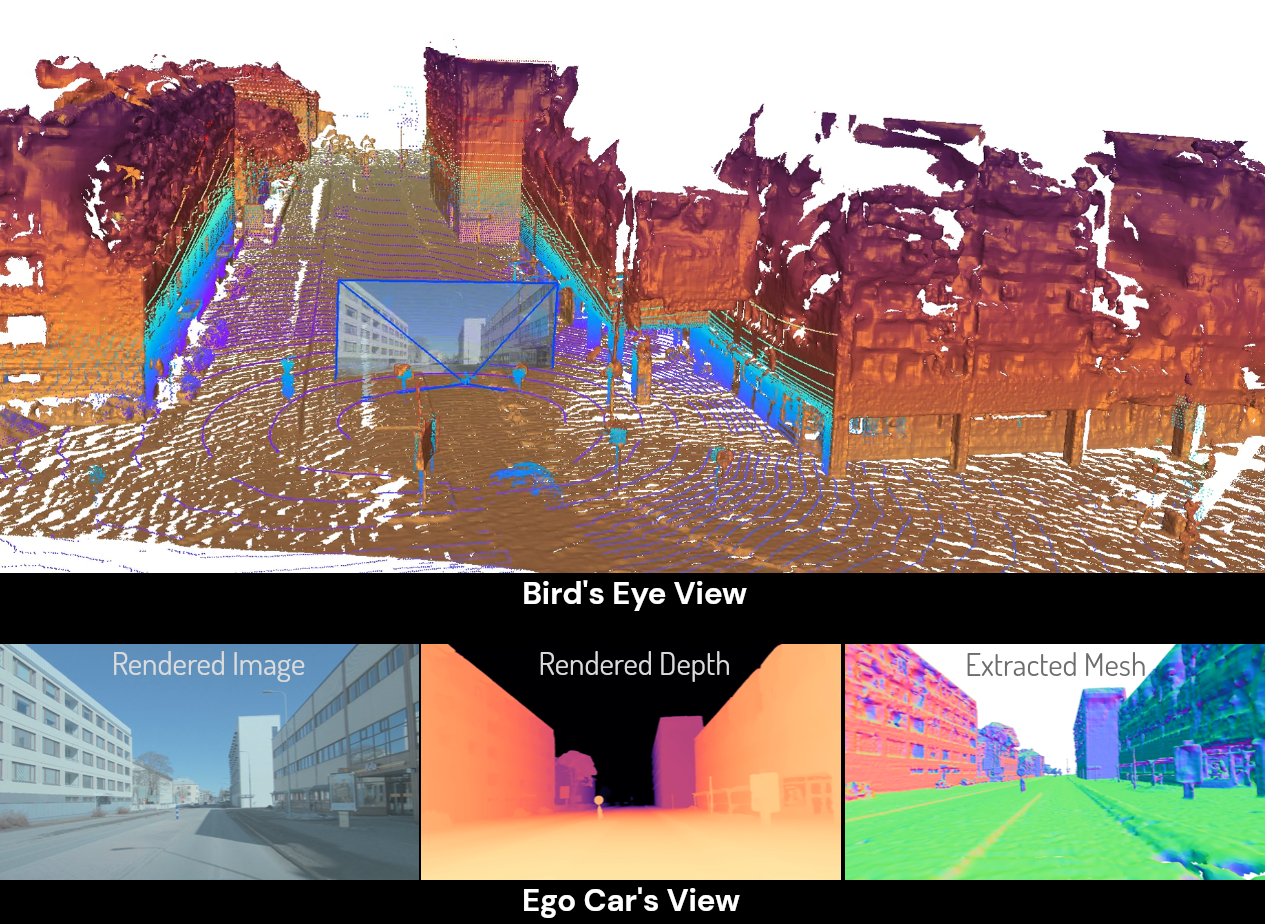}
\vspace*{-2mm}
\caption{We demonstrate dense and accurate 3D structure being represented by an implicit model, which can be further rendered into RGB images and depth maps or reconstructed into a high-quality mesh.}
\label{fig:intro}
\end{figure}

Reconstructing 3D urban environments from sensor data is an important problem with applications in autonomous vehicles, augmented reality, city planning and more. Traditional approaches rely on fusing data from LiDAR and structure from motion techniques applied to camera images. However, these explicit representations have limitations such as sparsity, noise, or difficulty scaling to large scenes. Neural implicit functions provide an alternative representation that is compact, smooth and can easily scale to model complex scenes. Recent works have shown promising results using neural radiance fields and occupancy networks to reconstruct indoor and small outdoor scenes from images~\cite{zhang2020nerf++,martin2021nerf,barron2022mip}. However, applying these techniques to reconstruct large-scale urban environments from vehicle sensors poses new challenges. In this paper, we introduce a method to urban scene reconstruction using a framework combining neural implicit surfaces and
radiance fields that addresses these challenges. Example use cases of the method include online 3D environmental model or offline extraction of 3D instances for multimodal data augmentation. In particular, we plan to deploy our trained model to aid our automated labelling pipeline.

%% file: include/method.tex
\begin{figure*}[t!]
    \centering
    \includegraphics[width = \textwidth]{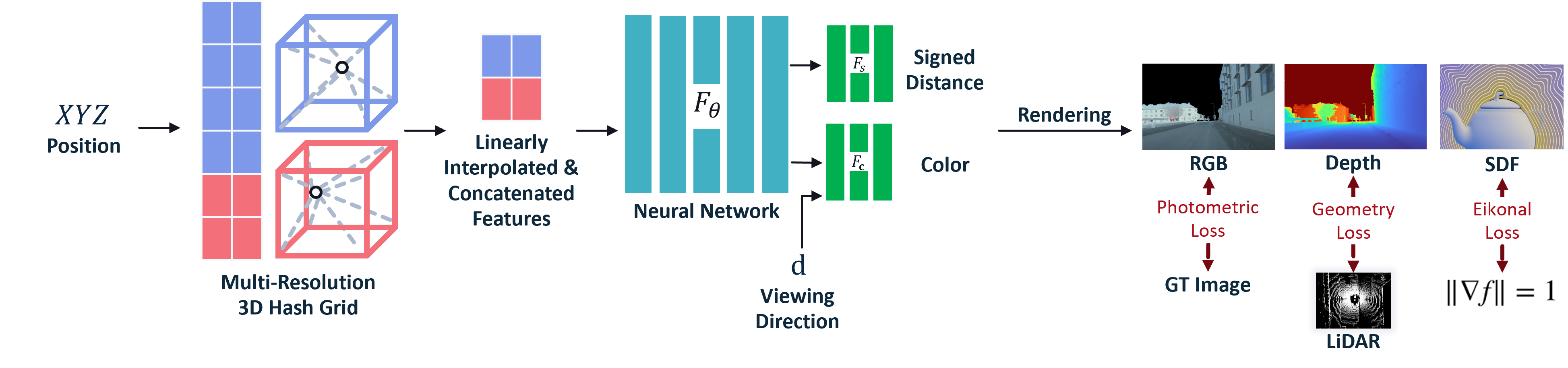}
    \caption{Overview of our foreground model. It follows the same multiresolution hash grid design as in~\cite{muller2022instant} and predicts signed distance rather than density. The other MLP head predicts view-dependent color by taking in the viewing direction. Three major supervisions are photometric loss to supervise the reconstructed scene appearance, Eikonal loss to regularize the learned SDF, and geometry loss to supervise the reconstructed scene geometry.}
    \label{fig:architect}
\end{figure*}

\section{METHODOLOGY} \label{sec:method}
In this section, we describe the 3D scene reconstruction framework, the challenges we have faced, as well as the corresponding solutions. 

\subsection{Background} \label{sec:background}

Neural Radiance Fields (NeRF)~\cite{mildenhall2021nerf} are composed of multiple MLPs that implicitly represent a target scene. A NeRF takes in a 3D position $\mathrm{x}$ and viewing direction $\mathrm{d}$, and outputs volume density $\sigma$ and color $c$. The viewing direction input enables NeRF to learn view-dependent color $c$ and hence to reproduce the reflective appearance of certain materials in the scene. On the other hand, the volume density output $\sigma$ is made viewing direction independent because intuitively volume density aims to reproduces the scene geometry, which is constant irrespective of viewing direction. 

To reproduce the appearance of the scene, or in other words, to render a pixel in an image, the renderer shoots a ray from the camera center $\mathrm{o}$ through the pixel to infinity, denoted as $\mathrm{r}(t)=\mathrm{o}+t\mathrm{d}$, where $\mathrm{d}$ is the direction from center camera to the pixel. Then it randomly samples points along the ray with distances $\{t_i\}_{i=0}^{N}$ from the $\mathrm{o}$. Those points $\mathrm{r}(t_i)$ and viewing direction $\mathrm{d}$ are passed to the MLPs which produce $\mathrm{c}_i$ and $\sigma_i$ for every point. Alpha compositing is used to get the final color of this pixel, with alphas equal to the predicted densities:
\begin{multline} \label{eq:render_color}
    \mathrm{c_{out}}=\sum_{i=1}^{N} T_i(1-e^{-\delta_i\sigma_i}) \mathrm{c}_i \\
    \text{where $T_i=\exp\left(-\sum_{j=0}^{i-1}\delta_j\sigma_j\right)$ and $\delta_i=t_i-t_{i-1}$}
\end{multline}
Similarly, to reproduce the geometry of the scene, or in other words, to render a pixel in a depth map, we simply replace $c_i$ in Equation~\ref{eq:render_color} with depth $t_i$:
\begin{equation} \label{eq:render_depth}
    \mathrm{d_{out}}=\sum_{i=1}^{N} T_i(1-e^{-\delta_i\sigma_i}) t_i
\end{equation}

Simple coordinates $\mathrm{x}=(X,Y,Z)$ lack the expressiveness for high-frequency details, so $\mathrm{x}$ and $\mathrm{d}$ are encoded into higher-dimensional vectors of sine and cosine functions, known as sinusoidal positional encoding~\cite{mildenhall2021nerf} $\gamma_{PE}$. 

However, naively sampling points along a ray becomes intractable in outdoor scenes because the spatial coordinate could theoretically be infinite (e.g., the sky). Therefore, NeRF++~\cite{zhang2020nerf++} decomposes the scene into foreground and background, with foreground enclosed by a volume of unit sphere and the background enclosed by a volume of an inverted sphere. Instead of inputting $\mathrm{x}$ into the background model, the input becomes $(\mathrm{x'}, 1/r)$ where $\mathrm{x'}$ is $\mathrm{x}$ projected onto the unit sphere and $r$ is the distance from $\mathrm{x}$ to the sphere's center, which effectively maps unbounded spatial inputs to bounded ones, avoids numeral unstable problems, and hence facilitates the representation of background scene.

After achieving high-quality representation, speed becomes a concern. Instant-NGP~\cite{muller2022instant} improves both training and rendering by using spatial data structures to store neural features which can be subsequently interpolated into feature vectors per spatial coordinate. It also adopts a multiresolution hash table for encodings, which significantly decreases the capacity required of the prediction MLPs. We refer to~\cite{muller2022instant} for details.

Although radiance field representations like a NeRF have shown great performance in novel view synthesis, they have been demonstrated to be unstable and ambiguous toward accurate geometry~\cite{wang2021neus,oechsle2021unisurf} because the underlying volume density is non-smooth and prone to artifacts. Intuitively, there's no constraint enforced on the volume density in empty space and hence it becomes difficult to extract watertight geometry from it. Therefore, prior work embraces neural implicit surface to achieve accurate geometry. Both VolSDF~\cite{yariv2021volume} and NeuS~\cite{wang2021neus} replace the output density $\sigma$ in NeRF by signed distance and then analytically convert the predicted signed distance to density to be used in the same way as volume rendering in NeRF. Because the network is now essentially a signed distance function (SDF), we can readily enforce the SDF prior in supervision, known as the Eikonal loss $\lvert \lVert \nabla f \lVert - 1 \rvert$. As such, the network represents the scene geometry with regularization, which is critical to our scene reconstruction use case. 

\subsection{Scene Decomposition}

Inspired by prior work~\cite{guo2023streetsurf,zhang2020nerf++,ost2021neural}, we decompose the scene into foreground and background and fit two different models to each of them. We define the foreground to be the street scene, including buildings, trees, road elements, etc., and define the background to be landscapes or the sky, regions that can't be reached by the ego vehicle in the current data sequence. We adapt our solution from StreetSurf~\cite{guo2023streetsurf} as well as use part of it as the backbone of our solution. 

Figure~\ref{fig:architect} illustrates the foreground model. Specifically, we follow the same design in~\cite{muller2022instant}, which stores feature embeddings in a multiresolution 3D hash grid and hence allows a much smaller and faster decoder than the standalone MLP in NeRF~\cite{mildenhall2021nerf}. Our network is an implicit surface model because besides the view-dependent color, it predicts the signed distance $s$ from the input position to the nearest surface, instead of predicting volume density as in NeRF. This helps extract watertight geometry from the learned model. We follow the analytical solution in NeuS~\cite{yariv2021volume} to convert the signed distance $s$ to density $\sigma$ in or to enable volume rendering for both color and depth:
\begin{equation}
    \sigma_i = \alpha \Phi_\beta\left(-s_i\right) 
\end{equation}
where $\Phi_\beta$ is the cumulative distribution function (CDF) of the Laplace distribution with zero mean and $\beta$ scale:
\begin{equation}
    \Phi_\beta(x) = 
    \begin{cases}
        \frac{1}{2}exp\left(\frac{x}{\beta}\right) & \text{if $x\leq 0$}\\
        1-\frac{1}{2}exp\left(-\frac{x}{\beta}\right) & \text{if $x>0$}
    \end{cases}
\end{equation}
Both $\alpha$ and $\beta$ are learnable parameters. Under this conversion, the density becomes $0$ when $s_i<0$ (i.e., outside the surface) and becomes $\alpha$ when $s_i\geq0$ (i.e., inside or on the surface), with the sharpness of drop of the CDF controlled by $1/\beta$. After approximating the density, we use Equation~\ref{eq:render_color} to render RGB color and Equation~\ref{eq:render_depth} to render depth. 

Our background model takes in the spatial coordinate $\mathrm{x}$ and applies the inverse sphere parameterization technique in NeRF++~\cite{zhang2020nerf++} to $\mathrm{x}$, so that the re-parameterized input becomes bounded even for landscapes and the sky at nearly infinite distances. 

The final rendering becomes the composite of foreground and background, equivalent to breaking the integral in Equation~\ref{eq:render_color} (or similarly Equation~\ref{eq:render_depth}) into two parts:
\begin{equation}
    \mathrm{c_{out}}=\sum_{i=1}^{N^{fg}} T_i(1-e^{-\delta_i\sigma_i}) \mathrm{c}_i + \sum_{i=1}^{N^{bg}} T_i(1-e^{-\delta_i\sigma_i}) \mathrm{c}_i
\end{equation} \label{eq:composite_render}
where $t_{N^{fg}}$ is both the farthest sampled depth in the foreground model and the closest sampled depth in the background model, and $t_{N^{bg}}=\infty$. 

\subsection{Dynamic Object Filtering}

We aim at 3D scene reconstruction to aid automated ground truth generation purposes and we focus on the static scene only. We filter independently moving objects in the scene based on annotations obtained from our 3D object detection (3DOD) pipeline which detects vehicles, pedestrians and two wheeled vehicles. Given the detections from the 3DOD pipeline, we can readily avoid any point being sampled by checking if the point is inside any bounding box during sampling and ray marching. To ensure that we use only the detections corresponding to dynamic objects, rather than transient objects that remain static in the current scene (e.g., parked vehicles, construction sites, etc.), we leverage two complementary signals: (1) the relative speed of the bounding box and (2) the difference in the absolute position of the bounding box across frames. The relative speed per bounding box is obtained by the 3DOD pipeline, which tells whether the annotated object is moving or not relative to the ego vehicle. The difference in the absolute position is obtained by transforming bounding boxes' positions to the world frame using ego vehicle's positions. If the relative speed is zero when the ego is moving, or the relative speed is nonzero when the vehicle is not moving, or the difference in the absolute position is nonzero, we keep the annotation for filtering; otherwise, we regard it as static objects. We demonstrate the effectiveness of dynamic object filtering in Section~\ref{sec:exp}. 

When we filter out dynamic objects during scene reconstruction, pose refinement becomes a well-posed problem as it would not be in dynamic environments where motion cues are not consistent between ego vehicle and dynamic objects~\cite{shen2023dytanvo}. Therefore, we set the input pose $\left[\mathrm{R} \vert t\right]$ to be a learnable parameter that is trained together with grid features and MLP weights. Due to difficulty of optimizing in $\mathrm{SO(3)}$, we parameterize the rotation component into unit quaternion $q$ and regress $\Delta q$.

\subsection{Large-Scale Support} \label{sec:large}

The size of the scene covered by each sequence varies, depending on the speed of the ego vehicle. However, the model capacity is predetermined. Therefore, we adopt a simple but effective 'divide and conquer' approach to larger-scale input sequences. With the same rationale as in Block-NeRF~\cite{tancik2022block}, we divide the input sequence into subsequences with a pre-fixed number of frames. Then we train our model on each subsequence in parallel as the reconstruction of each is independent. All subsequences still share the same world coordinate so that during rendering, we can readily merge all subsequences based on ego vehicle's position and orientation. 

\begin{figure}[b]
    \centering
    \includegraphics[width = \columnwidth]{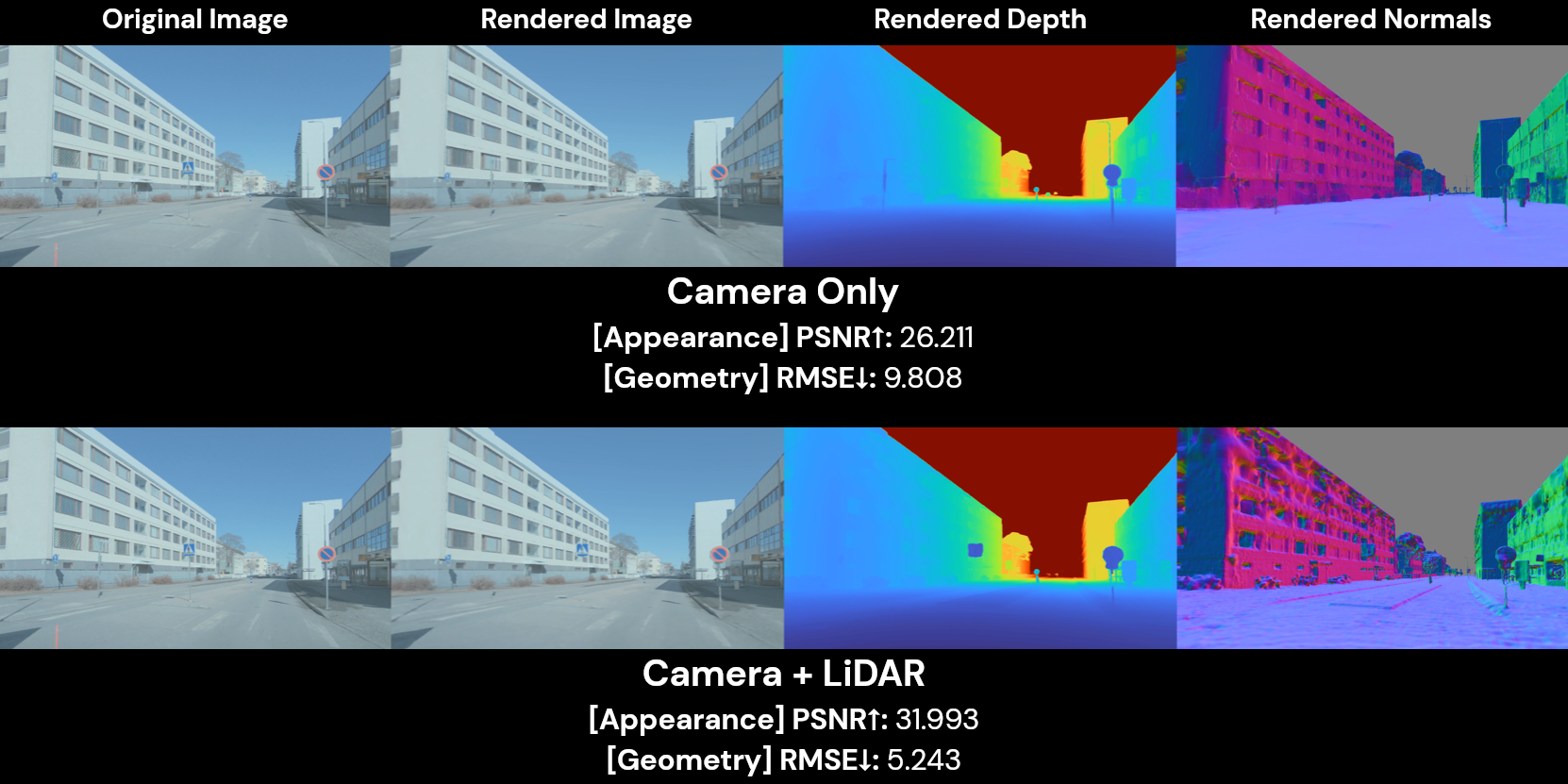}
    \vspace*{-3mm}
    \caption{Qualitative and quantitative results that demonstrate benefits of combining LiDAR and camera. PSNR$\uparrow$: $26.211$ without LiDAR and $31.993$ with LiDAR. RMSE$\downarrow$: $9.808$ without LiDAR and $5.243$ with LiDAR.}
    \label{fig:comb}
\end{figure}

\subsection{Supervision} \label{sec:sup}

In addition to the flow, Figure~\ref{fig:architect} also shows three major supervision signals (only for the foreground model for simplicity). To reconstruct appearance of the scene, we calculate the photometric loss $L_C$ between the full rendered image (i.e., after compositing both foreground and background) and input image. To reconstruct geometry of the scene, we first regularize the learned SDF field by the Eikonal loss $L_S$ (as introduced in Section~\ref{sec:background}), and then make use of the LiDAR measurement to calculate the geometry loss $L_D$. Point clouds are first projected onto the camera frame and compared against the rendered depth at valid coordinates. The final loss becomes $L=L_{C} + L_{S} + L_{D}$.

%% file: include/experiment.tex
\section{EXPERIMENTAL RESULTS}  \label{sec:exp}

We train and evaluate our method on the challenging internal automotive dataset. We train both foreground and background models jointly for 10,000 iterations with 8196 rays per batch. We use Adam~\cite{kingma2014adam} with a learning rate of $1\times10^{-2}$. Without large-scale support, we train one sequence on an NVIDIA Tesla V100, with large-scale support we parallelize training using multiple cards. 

We first demonstrate the reconstruction result in both bird's eye view (BEV) and ego car's view in Figure~\ref{fig:intro}. The complete video demonstration has been presented at the Electronic Imaging Autonomous Vehicles and Machines conference. In BEV, we extract the foreground mesh from the learned SDF field by the marching cubes algorithm~\cite{lorensen1998marching}, and further apply occlusion culling to it due to the noise in unobserved or occluded regions, such as the rear of the buildings. In the video, we overlay input image at each time and point clouds on top of the mesh. In the ego car's view, we show the rendered image, the rendered depth, and the extracted mesh colored by surface normals.

\begin{figure}[b]
    \centering
    \includegraphics[width = \columnwidth]{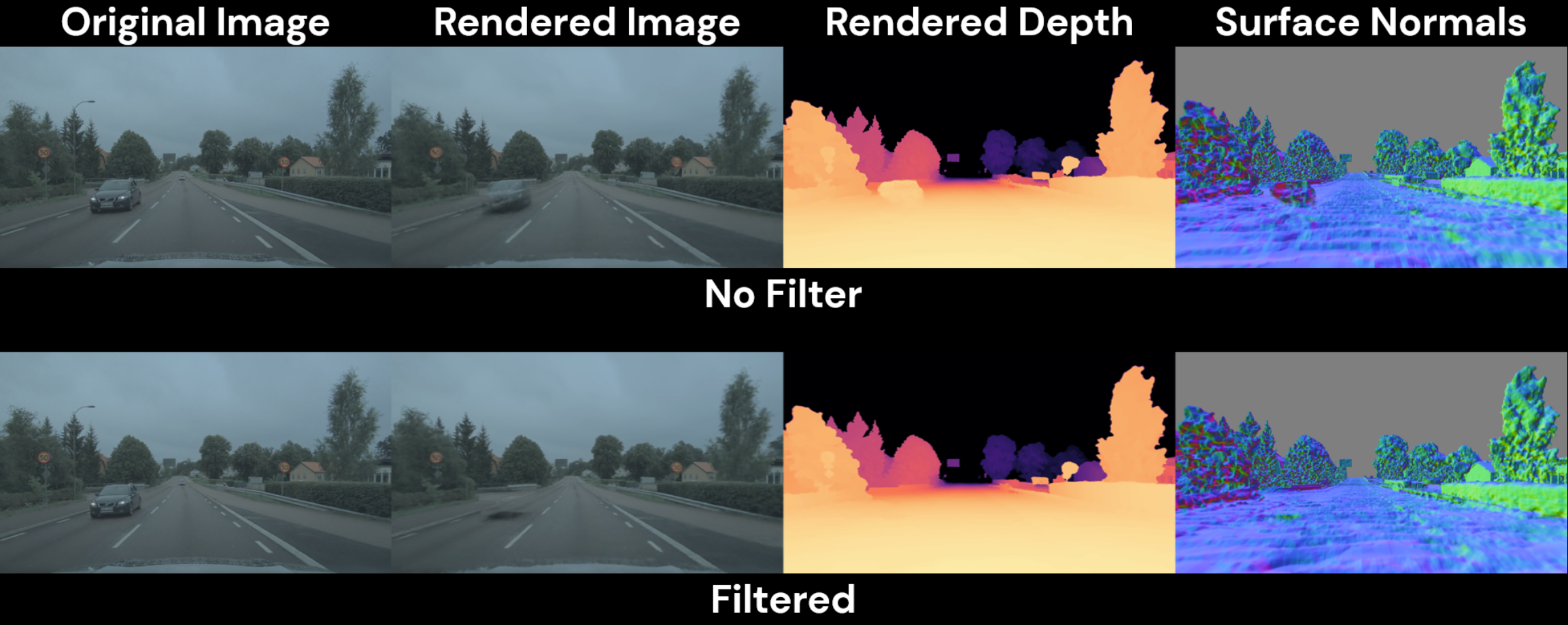}
    \vspace*{-3mm}
    \caption{Qualitative results of dynamic object filtering}
    \label{fig:dyna}
\end{figure}


Figure~\ref{fig:comb} illustrates qualitative and quantitative results of combining LiDAR and camera in neural rendering for urban scene reconstruction. Peak Signal-to-Noise Ratio (PSNR) and Root Mean Square Error (RMSE) are used to evaluate appearance reconstruction and geometry reconstruction respectively. Both metrics are calculated and averaged over the entire 300-frame sequence. In the camera-only baseline, we follow~\cite{guo2023streetsurf}, using off-the-shelf models~\cite{eftekhar2021omnidata} to estimate depth and surface normal as pseudo ground truth in depth supervision. Quantitatively, we see there’s a 23\% increase in PSNR and 46\% decrease in RMSE when we make use of LiDAR in supervision. Qualitatively, we see the camera-only baseline has missed one of the road signs in all three rendered modalities. Since LiDAR sweeps provide much more reliable depth even on tiny objects, the model is able to capture finer details. On the other hand, off-the-shelf monocular depth estimators still take camera data as input and hence the camera-only model can't recover accurate geometry. Note that the windows on the building are also distinguishable after using LiDAR, because monocular depth estimator can't perceive such detailed geometry cues. We have observed that the surface normals rendered by the camera+LiDAR model are much noisier than the camera-only one, especially on the building. This is because unlike per-pixel depth estimated by the monocular depth estimator, point clouds from LiDAR are much sparser and hence not every rendered depth pixel can be supervised by $L_D$ in Section~\ref{sec:sup}, which causes ambiguity in the surface. Further regularization losses can be applied to smoothen the surface in regions with undefined depth.

Figure~\ref{fig:dyna} illustrates the effectiveness of our dynamic object filtering implementation based on 3DOD detections. Without filter, the dynamic object appears as ghosting artifacts in the rendered image. And it's observable in both rendered depth and surface normals, which is challenging for accurate 3D scene reconstruction. By avoiding sampling inside the 3D bounding box during ray marching, the dynamic object becomes completely "invisible" in rendered depth and surface normals. Artifacts also disappear in render image, except that the shadow of the vehicle still remains on the ground because it's not enclosed in the annotation. Some recent works solve this by having an additional prediction head to predict a shadow ratio, and we consider it as a topic to be explored in the future. 

\begin{figure}[t]
\centering
\begin{subfigure}[b]{\columnwidth}
   \includegraphics[width=1\linewidth]{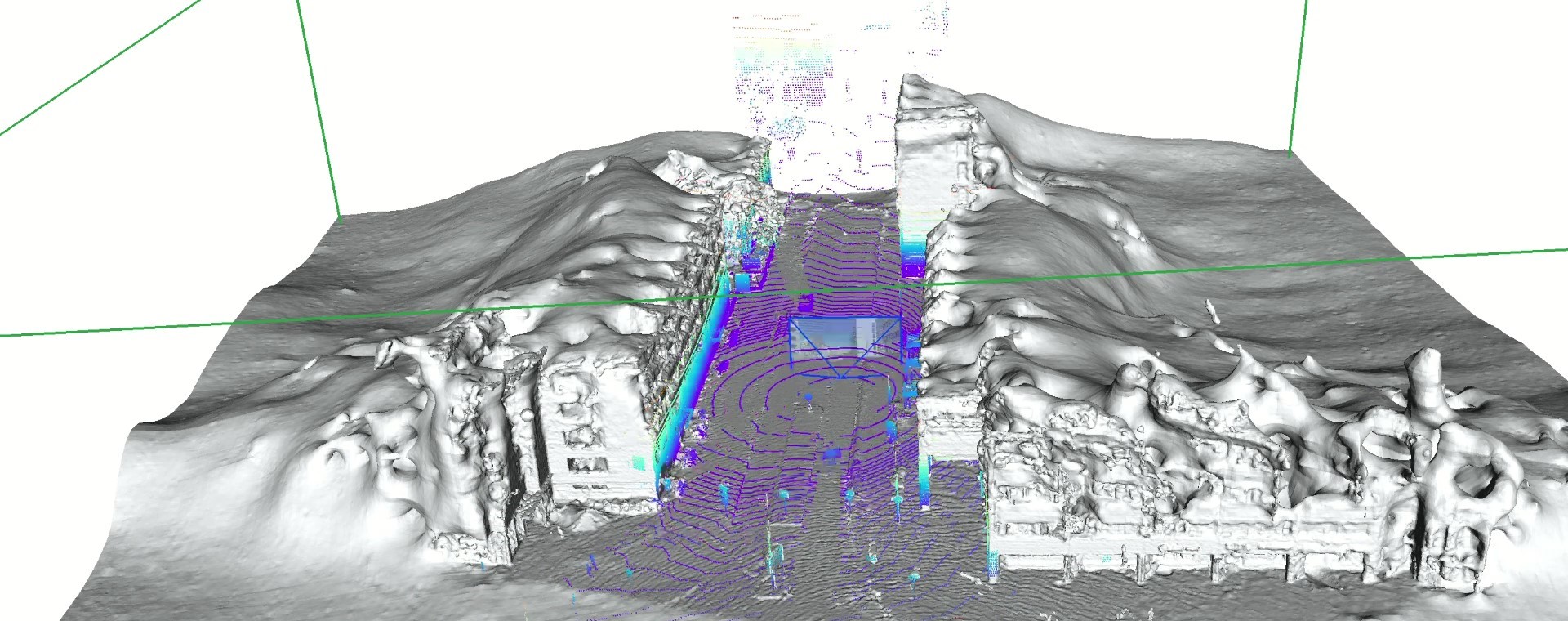}
\end{subfigure}
\begin{subfigure}[b]{\columnwidth}
   \includegraphics[width=1\linewidth]{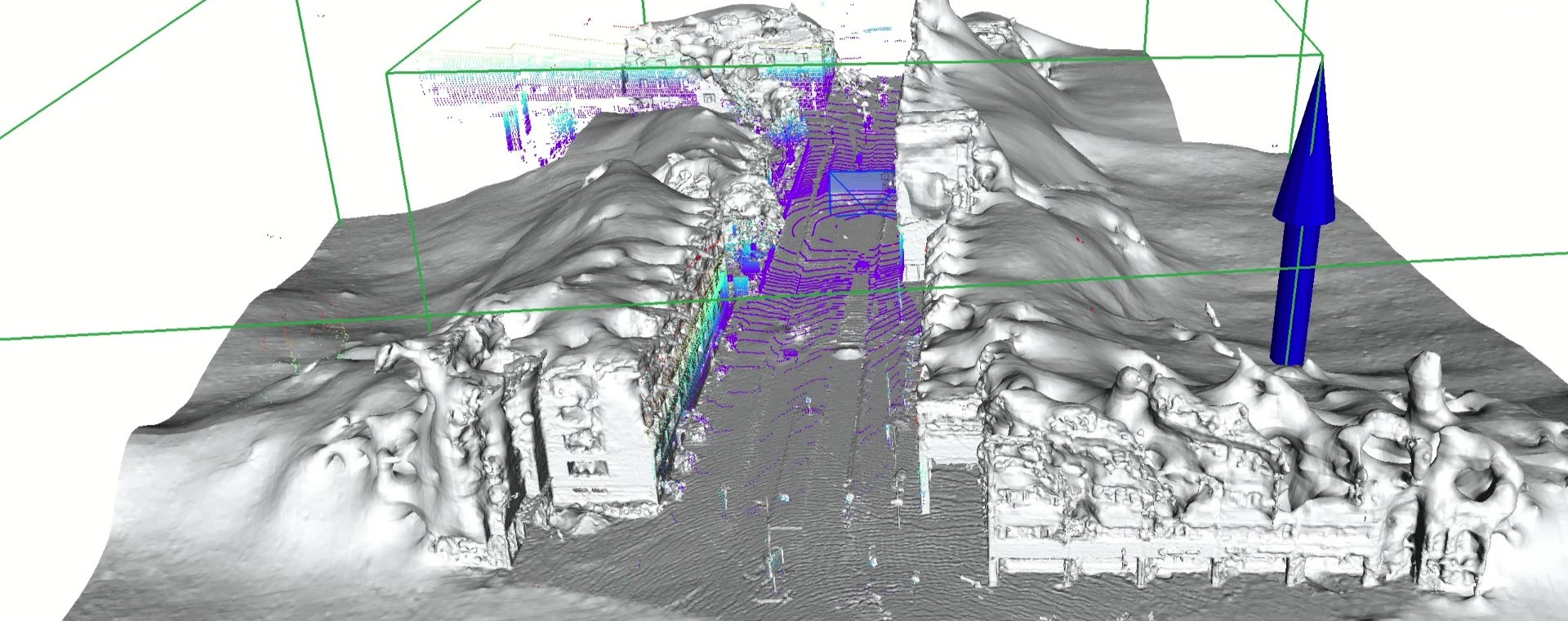}
\end{subfigure}
\caption{Large-scale support demonstration in BEV. Occlusion culling is not applied to the mesh for simplicity. Green boxes denote allocated spatial size for each subsequence. Top: Extracted mesh of the first subsequence. Bottom: Extracted mesh of the next subsequence and merged into the first subsequence.}
\label{fig:large}
\end{figure}

Finally we demonstrate the scalability of our solution to larger scenes. As detailed in Section~\ref{sec:large}, the sequence shown in Figure~\ref{fig:large} has been divided into two subsequences and two models are trained on each subsequence independently. Since they share the same coordinate, all types of renderings can be merged seamlessly. Trained models for each subsequence are queried sequentially according to the input position. Meshes are extracted independently from learned SDF fields and then merged together as shown in the bottom figure. 

%% file: include/conclusions.tex
\section{Conclusion} 

In this paper, we introduce our solution to large-scale urban scene reconstruction based on neural rendering. We demonstrate the method's efficacy in reliable neural scene representation by combining neural radiance fields and neural implicit surface. By leveraging LiDAR measurements, fine details in the scene can be captured and reconstructed accurately. Furthermore, we show the method is immune to disturbance of dynamic objects by leverage 3D object detections. We also prove the scalability of the method by reconstructing arbitrarily large environments through divide and conquer. In future work, we aim to to integrate our solution into our offline perception automated driving stack.

%% file: include/bio.tex
 \begin{biography}

\noindent \textbf{Shihao Shen} is an machine learning engineer at Qualcomm. He received B.S. in Electrical Engineering from the University of California San Diego and M.S. in Robotic Systems Development from Carnegie Mellon University. His main research focus is machine learning with applications in geometry vision, localization, and 3D reconstruction. \\

\noindent \textbf{Louis Kerofsky} is a researcher in video compression, video processing and display.  He received M.S. and Ph.D. degrees in Mathematics from the University of Illinois, Urbana-Champaign (UIUC).  He has over 20 years of experience in research and algorithm development and standardization of video compression.  He has served as an expert in the ITU and ISO video compression standards committees. He is an author of over 40 publications which have over 5000 citations.  He is an inventor on over 140 issued US patents.  He is a senior member of IEEE, member of Society for Information Display, and member of Association for Computing Machinery.  \\

\noindent \textbf{Varun Ravi Kumar} holds a staff engineer position in Qualcomm and leads the multi-modal perception team. He received a Ph.D. degree in Artificial Intelligence from TU Ilmenau in 2021 and an M.Sc. degree in 2017 from TU Chemnitz, Germany. Ph.D. thesis builds a first-ever 6-task multi-task learning near-field perception system that constitutes the necessary modules for a Level3 autonomous stack using surround-view fisheye cameras. He has 8+ years of experience in research focusing on designing self-supervised perception algorithms using neural networks for self-driving cars. He is an author of 27 publications with 900 citations and 80+ filed patents. \\

\noindent \textbf{Senthil Yogamani} holds an engineering director position at Qualcomm and leads the data-centric AI for autonomous driving department. He has over 18 years of experience in computer vision and machine learning including 15 years of experience in industrial automotive systems. He is an author of 120 publications which have 6500 citations and 150+ filed patents. He serves on the editorial board of various leading IEEE automotive conferences including ITSC and IV.



 \end{biography}